\documentclass[11pt,twocolumn]{article}

\usepackage[margin=1in]{geometry}
\usepackage{booktabs}
\usepackage{graphicx}
\usepackage{subcaption}
\usepackage{xcolor}
\usepackage{amsmath}
\usepackage{natbib}
\usepackage{url}
\usepackage{hyperref}
\usepackage{authblk}

\begin{document}


\title{Cross-Lingual Reusability of LLM Pipelines: Financial Extraction, Multilingual Reasoning, and Semantic Screening}

\author[1]{Joong Ho (Joshua) Choi}
\author[1]{Sherine George}
\author[1]{Jiayang Zhao}
\author[1]{Jayesh Rathi}
\author[2]{Ashwin Kandath}
\author[1]{Ananya Sane}
\author[1]{Vishvak Srinivasan}
\affil[1]{BNY AI Hub}
\affil[2]{BNY Responsible AI Office}

\maketitle


\begin{abstract}
Financial institutions increasingly deploy large language models (LLMs) in
multilingual markets, raising a practical question: can pipelines developed
and validated in English be \emph{reused} across languages, or must they be
rebuilt per locale?  We test cross-lingual reusability along three
dimensions---accuracy, reasoning-effort transfer, and
behavioural stability---through three complementary tasks.
\textbf{Task~1} (cheque extraction) and \textbf{Task~2} (Mintaka QA with
pinned evidence across nine languages and nine model--effort
configurations) test whether accuracy and effort settings transfer;
\textbf{Task~3} (FinanceBench pilot, 50 items over SEC~10-K filings in
four languages) validates whether these patterns generalise to financial
content.
\emph{Accuracy reusability} is broadly confirmed: Gemini~3.5 Flash achieves
95.2--97.6\% extraction accuracy across prompt languages, and
FinanceBench per-language means span 5.8pp (77.1--82.9\%), though the
language-difficulty ranking reverses between domains (Hindi is worst on
Mintaka, best on FinanceBench at 82.9\%).
\emph{Effort reusability} is domain-dependent: higher reasoning effort
yields marginal gains on general-knowledge QA but up to 8pp on
financial content, and effort configurations that help in English help
proportionally in other languages.
\emph{Behavioural reusability} holds: citation-set overlap with English
averages 77--79\% Jaccard, with no single language consistently worse.
Together, these results support qualified reuse of English-built
pipelines, but per-domain, per-language validation remains
necessary---neither semantic similarity nor self-reported confidence
reliably predicts cross-lingual accuracy.
\end{abstract}


\section{Introduction}
\label{sec:intro}
Financial institutions are rapidly adopting large language models (LLMs) for tasks ranging from document processing to knowledge retrieval. As these systems are deployed globally, a key question is whether solutions developed and validated in English—such as prompts, evaluation pipelines, and evidence corpora—transfer reliably to other languages. If translation alone preserves performance, institutions can reuse AI pipelines across markets rather than redeveloping them language by language.

We investigate this question through three tasks. \textbf{Task~1} examines document information extraction in a financial setting: given a handwritten cheque image, an LLM extracts the numeric courtesy amount and the legal amount written in words under prompts in different languages. \textbf{Task~2} examines cross-lingual question answering with deep research, where an LLM reasons over retrieved evidence to produce an answer. We evaluate this setting using the nine-language Mintaka benchmark~\citep{sen2022mintaka} with a fixed evidence corpus, allowing language and reasoning effects to be studied independently of retrieval variance.
\textbf{Task~3} then validates whether these patterns hold on
domain-specific financial content through a pilot study on the
FinanceBench benchmark~\cite{islam2023financebench}, evaluating 50
questions over SEC~10-K filings in four languages and directly testing
the generalisability of the RQ1--RQ2 findings beyond general-knowledge
content.

These three tasks are deliberately chosen to span distinct dimensions
of the LLM pipeline architectures that financial institutions deploy.
Task~1 is \emph{perception-dominated}: the
answer is visible on the cheque image, and the challenge is accurate
extraction under varied instruction languages---a low-reasoning,
high-accuracy setting typical of back-office operations (cheque clearing,
payment processing) where errors carry direct financial cost.  Task~2 is
\emph{reasoning-dominated}: the model must plan, synthesise, and judge
across multi-source evidence---a high-reasoning setting representative
of front-office knowledge work (research, compliance, due diligence).
Task~3 is \emph{domain-validation-focused}: it applies the same
cross-lingual QA design to financial filings, testing whether patterns
observed on general-knowledge questions extend to domain-specific
financial content.  The three tasks further differ in modality
(vision+text vs.\ text-only), in how language enters (instruction-only
vs.\ full cross-lingual context including translated evidence), in domain,
and in evaluation objectivity (exact numeric match vs.\ rule+judge
grading).  By holding model families constant across all three tasks, we
can test whether findings---such as
``model choice dominates language''---generalise across differences in
difficulty, domain, and modality rather than being artefacts of a single
task type.

Concretely, we test cross-lingual pipeline reusability along three
dimensions, each operationalised as a research question:

\begin{description}
  \item[\textbf{RQ1} (Accuracy reusability)] \emph{Can an
    English-validated LLM pipeline be reused in other languages without
    significant accuracy loss?}
    We test this on cheque extraction (Task~1), general-knowledge QA
    (Task~2), and financial QA (Task~3) across up to nine languages.
  \item[\textbf{RQ2} (Effort reusability)] \emph{Do reasoning-effort
    configurations tuned in English transfer to non-English settings?}
    We test this using 400 items $\times$ 9 languages $\times$ 9
    model--effort configurations with a pinned evidence corpus that
    eliminates retrieval variance (Task~2), plus a financial QA pilot
    (Task~3).
  \item[\textbf{RQ3} (Behavioural reusability)] \emph{Does citation and
    source-selection behaviour remain stable when the prompt language
    changes?}
    We measure citation-set overlap and proxy precision/recall across
    languages under fixed evidence (Task~2).
\end{description}


\section{Related Work}
\label{sec:related}

Several benchmarks evaluate LLM capabilities on financial tasks.
FLUE~\citep{shah2022flue} provides five English financial NLP
tasks---sentiment analysis, named entity recognition, structure boundary
detection, question answering, and stock movement prediction---and
introduces FLANG, a pre-trained language model for finance.
BloombergGPT~\citep{wu2023bloomberggpt} trains a 50B-parameter model on a
mixed finance--general corpus and evaluates it on standard financial NLP
benchmarks, demonstrating that domain-specific pre-training improves
financial task performance without degrading general capabilities.
FinBen~\citep{xie2024finben} offers a holistic evaluation across 35
datasets spanning seven task categories including information extraction,
question answering, and risk management.
FinanceBench~\citep{islam2023financebench} focuses on open-book financial
QA over public filings, testing retrieval-augmented generation in a
setting closer to real analyst workflows.  Despite their breadth, these
benchmarks are English-only or English-dominant.  Recent work has begun
to address this gap: MultiFinBen~\citep{peng2026multifinben} introduces a
multilingual financial benchmark across five languages, but evaluates
models on static task sets rather than testing whether an English-built
pipeline can be reused cross-lingually.

In document understanding, OCR-free vision--language models such as
Donut~\citep{kim2022donut} have enabled end-to-end extraction and have
been applied to cheque processing~\citep{zhao2024gencheck}.  Multimodal
LLMs also perform strongly on printed
documents~\citep{ma2024mmlongbench,pfitzmann2022doclaynet}, but
cross-lingual prompting effects on handwritten financial document
extraction remain largely unexplored~\citep{mondshine2025beyondenglish}.

Cross-lingual performance gaps in LLMs are well documented but poorly
explained.  \citet{sapenov2026mmpisabench} evaluate two proprietary LLMs
across 43 languages and show that token usage varies systematically by
language---some languages are simultaneously more expensive and less
accurate---suggesting that generation behaviour is not language-neutral
even when accuracy is controlled.
\citet{chen2025languageofthought} find that non-English reasoning traces
produce higher output diversity, with diversity increasing as the thinking
language diverges from English in embedding space.
\citet{wang2025explang} propose on-policy thinking-language selection and
show that multilingual chain-of-thought expands the exploration space,
achieving higher pass@$k$ under a fixed training budget.  Most relevant
to our work, \citet{lu2026beyondmonolingual} evaluate cross-lingual deep
research agents and find substantial degradation when evidence is
translated: accuracy remains lower even when all gold evidence is supplied
directly.  Deep research agents combine retrieval, planning, and
multi-step reasoning into end-to-end workflows, and
\citet{wang2026reflect} propose REFLECT for meta-evaluating such agents,
finding that even the best judges remain below 55\% accuracy on evidence
verification.  Crucially, because \citeauthor{lu2026beyondmonolingual}'s
benchmark does not control retrieval---each run searches anew---it cannot
fully separate retrieval degradation from reasoning degradation.

Translation quality is itself a potential confound in multilingual
evaluation.  \citet{thellmann2026translationerrors} find that target-side
translation errors are associated with 6--8 percentage-point reductions in
accuracy, increasing to 6--11 points among items answered correctly in
English, while automated error-span detection achieves only moderate
agreement with expert annotations.  These findings motivate interpreting
differences under translated inputs as end-to-end outcomes of a
translation-mediated pipeline rather than pure measures of intrinsic
language capability.

The effect of reasoning depth on LLM accuracy has been studied primarily
in English.  \citet{qin2025multilingualcot} provide a comprehensive
multilingual evaluation and find that (i)~thinking language affects
accuracy, (ii)~reasoning traces in different languages are not simple
translations, and (iii)~prompt language affects the final answer even when
the question is fixed.  \citet{deng2026thoughtsdepth} show that
low-resource languages degrade disproportionately as chain-of-thought
length increases, suggesting a resource-dependent ceiling on the benefit
of deeper reasoning.  \citet{gao2026languagelatent} frame language as a
latent variable in reasoning and argue that English as a default thinking
language is a root cause of cross-lingual bias.  These studies all
manipulate prompt-induced chain-of-thought; none examines the
\emph{provider-exposed} reasoning-effort controls---model-id suffixes,
thinking budgets---that production systems actually offer.

Finally, LLM-based evaluation itself introduces methodological risk.
\citet{haldar2025ratingroulette} find intra-rater
Krippendorff's~$\alpha$~\cite{krippendorff2011alpha} below the 0.80
threshold on multiple benchmarks.
\citet{chen2025pairwisebias} show that pairwise evaluation amplifies
superficial preferences; \citet{ye2024prepair} mitigate this with PRePair,
a pointwise-then-compare protocol.  In the multilingual setting,
\citet{hada2025multilingualjudge} report cross-language
$\kappa \approx 0.3$, and \citet{lee2026fairnessfluency} show that answer
language has more impact on judge bias than question language.

Our work addresses these gaps jointly.  We systematically vary prompt
language across a financial extraction task and a general-knowledge
reasoning task, covering three model families, and introduce
a pinned-evidence methodology that isolates reasoning quality from
retrieval quality in cross-lingual evaluation.  We study provider-exposed
reasoning-effort controls as deployed in production systems, where the
mechanism is opaque and family-specific.  To mitigate judge unreliability,
we use deterministic rule-based grading as the primary path---reserving the
LLM judge for cases the rules cannot settle.  Comparing
provider-controlled retrieval in our native deep-research track against
the shared-evidence track reveals how much accuracy depends on retrieval
versus reasoning.


\section{Data}
\label{sec:data}

\subsection{Handwritten Cheque VQA Dataset (Task~1)}
\label{sec:data:cheque}

\begin{figure}[t]
  \centering
  \includegraphics[width=\columnwidth]{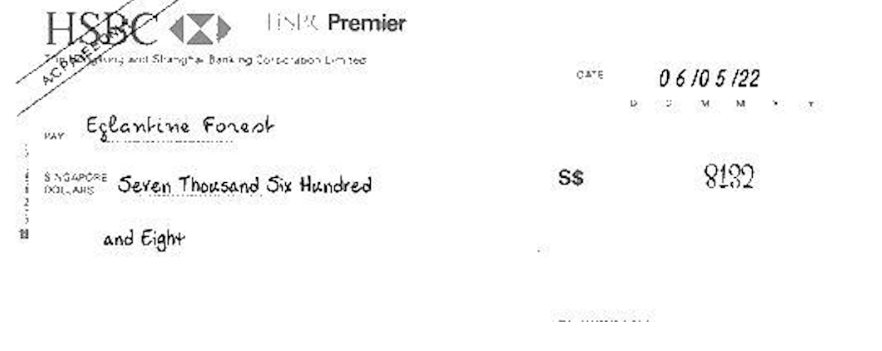}
  \caption{Example cheque image from the Handwritten Cheque VQA
    dataset~\cite{verma2024chequeVQA}.  The model must extract the
    courtesy amount (numeric, from the amount box) and the legal amount
    (written in words, converted to a number).  All cheque images are in
    English; only the prompt language varies across experimental
    conditions.}
  \label{fig:cheque-example}
\end{figure}

For Task~1 we use the Handwritten Cheque VQA
dataset~\cite{verma2024chequeVQA}, containing 3,312 question--answer pairs
over cheque images (Figure~\ref{fig:cheque-example}).  Each item provides
the amount in words (\emph{legal amount}) and in figures (\emph{courtesy
amount}).  We apply three successive filters: retain only items where both
amounts are annotated (914), then only items where the two amounts differ
so the model cannot copy one to the other (252), and finally only items
where both fields parse to numeric values for exact-match evaluation (124).

Each of the 124 items is evaluated across three model families and five
prompt languages (1,860 API calls total).  All cheque images are in
English; the multilingual dimension enters only through the prompt
language.

\subsection{Mintaka Benchmark (Task~2)}

Mintaka~\cite{sen2022mintaka} is a multilingual question-answering dataset
built on Wikidata.  Each of its 14,000 items contains a question in English
plus translations into eight languages (Arabic, German, Spanish, French,
Hindi, Italian, Japanese, Portuguese), typed answers (entity, numerical,
boolean, date, or string), eight topic categories (e.g.\ geography, movies,
sports), and nine complexity types including multihop, superlative, and
comparative questions.  This combination of nine-language coverage and
verifiable answers makes it well-suited for controlled cross-lingual
evaluation.

Although Mintaka's content spans general knowledge rather than finance, its
value here is as a controlled testbed for the \emph{pipeline architecture}
that financial institutions deploy: retrieval, multi-step reasoning, and
report synthesis over a multilingual question set with verifiable answers.
Existing multilingual financial QA benchmarks either lack the nine-language
breadth needed for our design (FinanceBench is English-only; MultiFinBen
covers five languages) or do not provide structured answers suitable for
automated evaluation at scale.  We draw a stratified random sample of 400
items (category~$\times$~complexity type) that underpins both evaluation
tracks.

\subsection{FinanceBench Pilot (Task~3: Domain Validation)}
\label{sec:data:financebench}

To test whether the cross-lingual patterns observed on Mintaka's
general-knowledge content hold on financial material, we conduct a
domain-validation pilot using FinanceBench~\cite{islam2023financebench}, an
open-source QA benchmark of 150 questions over SEC~10-K filings from 32
publicly traded companies.  We draw a stratified sample of 50 items
(${\sim}17$ each of \emph{metrics-generated}, \emph{domain-relevant}, and
\emph{novel-generated} questions) and evaluate them in four languages
(English, Hindi, Spanish, German) across the same three model families
and all nine model--effort configurations.  Evidence passages from the
source 10-K filings are translated into each target language using the
same GPT-4.1 translator~\citep{openai2025gpt41} (with Claude
Sonnet~4.5~\citep{anthropic2025sonnet45} as a fallback) employed for
the Mintaka evidence corpus, and questions and gold answers are
translated identically.  The pilot yields $50 \times 4 \times 9 = 1{,}800$
API calls---a deliberately moderate footprint designed to confirm or
refute the Mintaka patterns on domain-specific financial content,
including the effect of reasoning effort, rather than to serve as a
comprehensive financial benchmark.


\section{Method}
\label{sec:method}

All three tasks are evaluated on the same three model families---
GPT-5.1~\citep{openai2025gpt51},
Claude Opus~4.6~\citep{anthropic2026opus46}, and
Gemini~3.5 Flash~\citep{google2026gemini35}---enabling cross-task comparison of
multilingual effects.  Task~1 tests a single reasoning-effort level per
family across five prompt languages; Task~2 varies reasoning effort
across three levels per family (nine model configurations in total)
and nine languages; Task~3 uses the same nine model--effort
configurations across four languages
(Section~\ref{sec:data:financebench}).

\subsection{Task~1: Cheque Amount Extraction}
\label{sec:method:cheque}

Each cheque image is sent to the model with a structured prompt requesting
JSON output containing two fields: \texttt{Courtesy Amount} (extracted from
the amount box as a float) and \texttt{Legal Amount} (converted from words
to a numeric float), along with confidence scores for each.  The system
prompt identifies the model as a cheque information extraction system.  The
model is instructed to extract each amount independently and not copy one to
the other.

To test whether prompt language affects extraction accuracy, we translate the
extraction prompt into four additional languages---Korean, Japanese, Hindi,
and Spanish---using GPT-4.1 at temperature~0, while keeping the cheque
images unchanged (all cheques are in English).  GPT-4.1 was chosen because
it is not among the evaluated models (GPT-5.1, Claude Opus~4.6,
Gemini~3.5 Flash),
avoiding circularity between the translator and the systems under test.
This isolates the effect of prompt language from document language.
Each of the 124 evaluation cheques is processed with all five prompt
languages under the same model.

We evaluate three model families through the same orchestration path described in
Section~\ref{sec:method:design} and report three accuracy measures: \emph{courtesy accuracy} (predicted
courtesy amount equals ground truth), \emph{legal accuracy} (predicted
legal amount equals ground truth), and \emph{joint accuracy} (both correct
simultaneously).  All comparisons use exact numeric match after
normalization.

\subsection{Task~2: Cross-Lingual QA Design}
\label{sec:method:design}

Our experiment is a fully crossed design over three axes: \emph{item} (400
Mintaka questions~\cite{sen2022mintaka}, stratified by category $\times$
complexity type), \emph{language} (9: en, ar, de, es, fr, hi, it, ja, pt),
and \emph{model configuration} (9, spanning three families---GPT-5.1,
Claude Opus~4.6, and Gemini~3.5 Flash---each at three reasoning-effort
levels).
To control for retrieval variance---repeat runs of the same query returned
different evidence in pilot tests---we construct a \emph{pinned evidence
corpus}: one English Google Custom Search per item, captured once and
committed to version control.  An offline extraction step then produces
attributable source passages (1,066 passages across 400 items, mean 2.7
per item), which every subsequent evaluation run reads verbatim.

Task~2 uses two evaluation tracks.  For reasoning evaluation, under \textbf{shared evidence}, each
question--language pair receives a language-matched translation of the pinned
passages.  Because the underlying evidence and its translations are pinned
across model configurations, cross-lingual differences cannot arise from
retrieval variation; they may nevertheless reflect both model reasoning and
translation characteristics.  Questions use Mintaka's human translations; evidence is translated
by GPT-4.1 (temperature~0) with a URL-preserving validation step and
Claude Sonnet~4.5 as a fallback---neither translator is among the evaluated
models. For retrieval evaluation, we also employ native deep research with OpenAI and Gemini 
to execute their own end-to-end retrieval-to-report pipelines, so comparing the two tracks isolates retrieval quality as a factor.

Grading is two-stage: script-aware rule-based matching first (boolean
lexicon, numeric comparison, normalised entity containment), then a GPT-5.2~\citep{openai2025gpt52}
LLM judge for the remaining items.

\subsection{Semantic-Similarity Diagnostic}
\label{sec:method:statistics}

As an additional semantic diagnostic, we embed each model answer and
its ground truth using OpenAI's \texttt{text-embedding-ada-002}
model~\citep{openai2022ada002} and MPNet~\citep{song2020mpnet} to
consider encoder-specific artefacts.  For each prediction we record the
cosine similarity
\begin{equation}
  S_{f,i,\ell} \;=\; \max_{g \in G_i}
    \frac{\mathbf{e}_{\text{pred}} \cdot \mathbf{e}_{g}}
         {\|\mathbf{e}_{\text{pred}}\|\;\|\mathbf{e}_{g}\|},
  \label{eq:sim}
\end{equation}
where $G_i$ is the set of accepted aliases for item~$i$.  This score is
a continuous semantic proximity measure, not a binary correctness label;
the analysis is restricted to entity answers.

We pair each target-language response with the English response for the
same item and model configuration.  The item-level
\emph{language--configuration interaction} is
\begin{equation}
  \begin{aligned}
    I_{f,i} ={}& \frac{1}{8}\sum_{\ell \ne \mathrm{en}}
      \Delta S_{f,i,\ell} \\
      &{}- \Delta S_{f,i,\mathrm{en}},
  \end{aligned}
  \label{eq:interaction}
\end{equation}
where $\Delta S$ is the low-to-high (or standard-to-adaptive) change in
cosine similarity.  A positive $I_{f,i}$ means the configuration change
narrows the cross-lingual gap.  We average over items and compute 95\%
percentile intervals from 5{,}000 bootstrap samples stratified by
category~$\times$~complexity type.

\subsection{Citation and Source-Selection Metrics}
\label{sec:method:citation-bias}

To assess whether models change how they use evidence across languages
(RQ3), we extract the numbered references each model cites and compare
citation behaviour between each target-language variant and its
English counterpart for the same item and model configuration.
Six metrics are computed:
\begin{itemize}
  \item \textbf{Cited rate}: fraction of answers containing at least one
    citation.
  \item \textbf{Citation coverage}: share of available references cited,
    adjusting for the variable number of references per item.
  \item \textbf{Blanket citation rate}: fraction of answers citing every
    available reference (i.e.\ no selective sourcing).
  \item \textbf{Proxy precision / recall}: overlap between cited
    references and \emph{proxy alias-coattributed references}---those
    co-attributed with extracted passages containing a known gold-answer
    alias.  Boolean items are excluded because their gold strings
    match ubiquitously.
  \item \textbf{Citation-set Jaccard similarity}: $|R_{\text{en}} \cap
    R_{\ell}| \,/\, |R_{\text{en}} \cup R_{\ell}|$ between the valid
    reference sets selected in English and the target language for the
    same item.
\end{itemize}

\noindent
Binary metrics are tested with exact paired McNemar tests; numerical
metrics use paired bootstrap confidence intervals and exact sign tests.
All $p$-values are adjusted within each metric using the
Benjamini--Hochberg procedure.  Because each non-English condition pairs
a translated question with translated evidence, observed differences are
\emph{localised-condition effects relative to English}, not pure
question-language effects.

\section{Results}
\label{sec:results}

\subsection{Task~1: Cheque Amount Extraction}
\label{sec:results:cheque}

Table~\ref{tab:cheque-results} reports extraction accuracy on the
124-item evaluation set.  English-prompt joint accuracy ranges from
49.2\% (GPT-5.1) to 83.9\% (Claude Opus~4.6) to 96.0\% (Gemini~3.5 Flash).
When the prompt is translated into four additional languages while
keeping the cheque images in English, the multilingual effect is
model-dependent.  Non-English prompts \emph{boost} GPT-5.1: Spanish
reaches 54.8\% joint accuracy (+5.6pp over English), and all four
languages match or exceed the English baseline.  Conversely, non-English
prompts \emph{degrade} Claude Opus~4.6: Korean drops to 76.6\% ($-$7.3pp).
Notably, the joint drop ($-$7.3pp) exceeds the sum of the individual
field drops (courtesy $-$4.0pp, legal $-$2.4pp), indicating that errors
which overlapped in English become spread across different items under
Korean prompting---a de-correlation effect that amplifies the joint
penalty.
Gemini~3.5 Flash is near-ceiling across all languages (95.2--97.6\% joint),
with a range of only 2.4pp.

Two findings emerge.  First, model family dominates
prompt language as a factor in extraction accuracy: Gemini~3.5 Flash's weakest
language (Korean, 95.2\%) exceeds Claude Opus~4.6's best (English, 83.9\%)
by 11.3pp, which in turn exceeds GPT-5.1's strongest (Spanish, 54.8\%) by
over 29 percentage points.  Second, the multilingual prompting effect is
model-dependent: it provides a substantial boost for GPT-5.1 (up to +5.6pp),
is negligible for Gemini~3.5 Flash ($\leq$2.4pp range), and actually \emph{hurts}
Claude Opus~4.6 (up to $-$7.3pp for Korean), suggesting that the direction and
magnitude of the multilingual effect cannot be assumed uniform across model
families.

\begin{table}[t]
  \centering
  \scriptsize
  \setlength{\tabcolsep}{2.5pt}
  \begin{tabular}{llccc}
    \toprule
    \textbf{Model} & \textbf{Prompt} & \textbf{Courtesy} & \textbf{Legal} & \textbf{Joint} \\
    \midrule
    GPT-5.1   & English & 63.7\% & 83.1\% & 49.2\% \\
              & Spanish & 70.2\% & 83.1\% & 54.8\% \\
              & Japanese & 65.3\% & 84.7\% & 54.0\% \\
              & Hindi   & 67.7\% & 83.9\% & 52.4\% \\
              & Korean  & 63.7\% & 84.7\% & 50.8\% \\
    \midrule
    Claude Opus~4.6 & English & 85.5\% & 97.6\% & 83.9\% \\
              & Korean  & 81.5\% & 95.2\% & 76.6\% \\
              & Spanish & 83.9\% & 97.6\% & 81.5\% \\
              & Hindi   & 84.7\% & 97.6\% & 82.3\% \\
              & Japanese & 83.9\% & 96.8\% & 80.6\% \\
    \midrule
    Gemini~3.5 Flash & English & 96.0\% & 100\%  & 96.0\% \\
              & Spanish & 97.6\% & 100\%  & 97.6\% \\
              & Japanese & 96.8\% & 100\%  & 96.8\% \\
              & Hindi   & 96.8\% & 100\%  & 96.8\% \\
              & Korean  & 95.2\% & 100\%  & 95.2\% \\
    \bottomrule
  \end{tabular}
  \caption{Cheque amount extraction accuracy by prompt language and model
    family on 124 evaluation items.  \emph{Courtesy}: numeric amount from
    the amount box; \emph{Legal}: amount in words converted to number;
    \emph{Joint}: both correct.  All cheque images are in English; only
    the prompt language varies.  Gemini achieves the highest baseline
    (96.0\% joint), Claude 83.9\%, and GPT 49.2\%.  The multilingual
    effect is model-dependent: GPT-5.1 benefits, Gemini is near-ceiling,
    and Claude shows mild degradation with non-English prompts.}
  \label{tab:cheque-results}
\end{table}

\subsection{Cross-Lingual QA Results across Tasks 2 and 3}

\label{sec:results:rules-accuracy}
\textbf{Accuracy is consistent cross-lingually, but language rankings are
domain-dependent.} On Mintaka, Claude Opus~4.6 leads at 79.5\%, followed by GPT-5.1 at
77.5\% and Gemini~3.5 Flash at 77.4\%; English is the strongest language
for every configuration (Table~\ref{tab:conditions}), but the
cross-lingual spread is modest---at most 3.3pp for GPT-5.1 low
(78.8\% vs.\ 75.5\% Japanese) and 3.0pp for Claude standard (80.2\%
vs.\ 77.2\% Japanese).
On FinanceBench, overall accuracy is 79.9\%  with the same
model-family ranking (Claude 81.8\%, GPT-5.1 80.3\%, Gemini 77.5\%) and
per-language means spanning 5.8pp (77.1--82.9\%).  However, the language
ranking \emph{reverses}; Hindi is the best non-English language (82.9\%),
outperforming even English (78.9\%), while German is the weakest (77.1\%).
A plausible explanation is that FinanceBench questions are shorter and
more formulaic than Mintaka's diverse trivia, reducing the surface area
for translation errors that disproportionately affect morphologically
rich languages.  The finding suggests that the ``Hindi penalty'' on
general-knowledge QA is \emph{domain-dependent} rather than an intrinsic
language limitation.

\textbf{Reasoning-effort gains are domain-dependent: marginal on general
knowledge, substantial on financial QA.}
On Mintaka, higher effort produces at most $+$0.8pp for GPT-5.1 and
Claude, and effectively zero for Gemini.  On FinanceBench, the effect is
larger but not uniform across families: GPT-5.1 jumps $+$7.0pp (76.0\%
to 84.0\% in English), Claude gains only $+$1.5pp, and Gemini
\emph{declines} $-$2.5pp.  The GPT-dominated lift is consistent with
numerical-reasoning demands of SEC~10-K questions benefiting from
additional compute in a way that trivia questions do not.

\begin{table*}[t]
  \centering
  \footnotesize
  \setlength{\tabcolsep}{3pt}
  \begin{tabular}{llrrrrrrrrrcrrrr}
    \toprule
    & &
    \multicolumn{9}{c}{\textbf{Task~2: Mintaka (9 languages)}} & &
    \multicolumn{4}{c}{\textbf{Task~3: FinanceBench}} \\
    \cmidrule(lr){3-11}\cmidrule(lr){13-16}
    \textbf{Family} & \textbf{Level} &
    \textbf{en} & \textbf{ar} & \textbf{de} & \textbf{es} &
    \textbf{fr} & \textbf{hi} & \textbf{it} & \textbf{ja} & \textbf{pt} &&
    \textbf{en} & \textbf{hi} & \textbf{es} & \textbf{de} \\
    \midrule
    GPT-5.1          & low       & 78.8 & 76.5 & 76.8 & 76.8 & 77.2 & 76.2 & 77.5 & 75.5 & 78.2 && 76.0 & 80.0 & 78.0 & 74.0 \\
                     & medium    & 78.5 & 77.0 & 77.2 & 77.8 & 77.9 & 77.5 & 77.8 & 76.2 & 78.0 && 78.0 & 82.0 & 82.0 & 78.0 \\
                     & high      & 79.0 & 77.2 & 77.2 & 78.2 & 77.7 & 77.8 & 78.2 & 77.0 & 77.8 && 84.0 & \textbf{88.0} & 84.0 & 80.0 \\
    \midrule
    Claude Opus~4.6  & standard  & 80.2 & 78.0 & 79.2 & 80.0 & 79.2 & 78.8 & 78.8 & 77.2 & 79.2 && 78.0 & 84.0 & 78.0 & 82.0 \\
                     & thinking  & 80.8 & 78.8 & \textbf{80.5} & \textbf{80.5} & \textbf{79.4} & \textbf{79.8} & \textbf{79.2} & \textbf{78.5} & \textbf{80.2} && 84.0 & 86.0 & \textbf{86.0} & 76.0 \\
                     & adaptive  & \textbf{81.0} & \textbf{79.5} & \textbf{80.8} & \textbf{80.5} & 79.2 & 79.5 & 79.0 & 78.2 & 79.5 && 84.0 & 84.0 & 82.0 & 78.0 \\
    \midrule
    Gemini~3.5 Flash & low       & 78.8 & 76.5 & 77.8 & 77.2 & 77.9 & 77.5 & 77.0 & 76.2 & 78.2 && 72.0 & 82.0 & 82.0 & 76.0 \\
                     & medium    & 78.5 & 76.0 & 77.8 & 77.5 & 77.7 & 77.2 & 76.8 & 77.2 & 77.2 && 80.0 & 82.0 & 76.0 & 78.0 \\
                     & high      & 78.2 & 76.0 & 77.8 & 77.2 & 77.9 & 77.2 & 77.8 & 77.5 & 77.5 && 74.0 & 78.0 & 78.0 & 72.0 \\
    \bottomrule
  \end{tabular}
  \caption{Cross-lingual QA accuracy (\%) by model configuration
    and language on Task~2 (Mintaka, 400 items per cell) and Task~3
    (FinanceBench, 50 items per cell).  On Mintaka, Claude at
    thinking/adaptive dominates; on FinanceBench, GPT-5.1 at high
    leads.  Hindi is the weakest language on Mintaka but the
    \emph{strongest} on FinanceBench, confirming that language
    difficulty is domain-dependent.
    \textbf{Bold} marks each column's maximum.}
  \label{tab:conditions}
\end{table*}

\begin{figure*}[ht]
  \centering
  \begingroup
  \setlength{\unitlength}{1pt}
  \setlength{\fboxsep}{1pt}
  \begin{picture}(469.755,166.926)
    \put(0,0){\includegraphics[width=\textwidth]
      {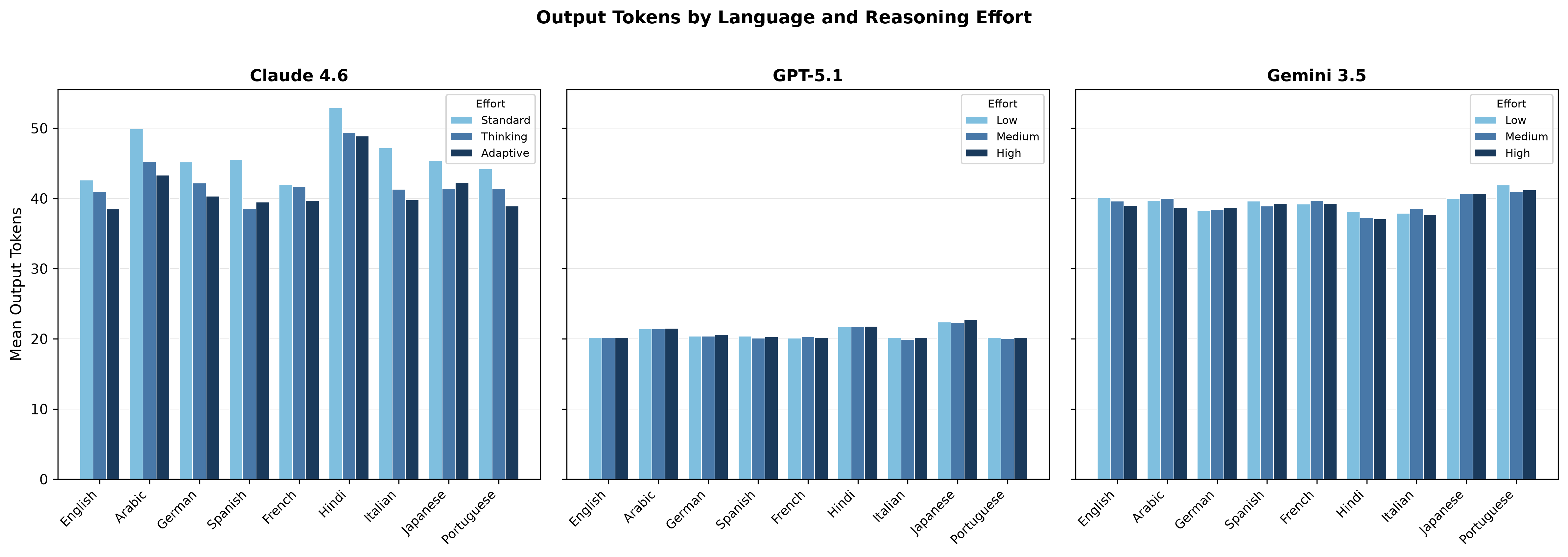}}
    \put(40,146){\colorbox{white}{\rule[-6pt]{0pt}{13pt}\makebox[100pt][c]{\fontsize{6}{7}\selectfont\bfseries Claude Opus 4.6}}}
    \put(340,146){\colorbox{white}{\rule[-6pt]{0pt}{13pt}\makebox[100pt][c]{\fontsize{6}{7}\selectfont\bfseries Gemini 3.5 Flash}}}
  \end{picture}
  \endgroup
  \caption{Mean output tokens by language and reasoning effort, counted with
    each family's native tokenizer (not comparable across panels).  Each
    panel shows one model family with bars grouped by effort level.
    GPT-5.1 has the smallest relative cross-language variation
    (${\sim}12\%$); Claude Opus~4.6 the largest ($>$20\%), with Hindi as the
    most expensive language and standard mode producing more tokens than
    thinking or adaptive modes.}
  \label{fig:lang_output}
\end{figure*}

\label{sec:results:output}
\textbf{Model selection dominates language as a cost driver, and the verbosity-as-incorrectness signal is domain-dependent.}
On Mintaka, reasoning effort modulates output length only for Claude
(${\sim}11\%$ reduction from standard to adaptive); GPT-5.1 and
Gemini~3.5 Flash are virtually invariant (Figure~\ref{fig:lang_output}).
Cross-language variation is likewise family-specific: GPT and Gemini are
compact (${\sim}10$--$12\%$ spread), whereas Claude's cross-language
range exceeds $20\%$, with Hindi as the most expensive language.  Incorrect answers are consistently
longer than correct ones---most markedly for Claude
($1.95{\times}$)---suggesting verbosity as a lightweight quality signal
(Figure~\ref{fig:outlen-correctness}).
On FinanceBench, however, this pattern reverses: incorrect answers are
\emph{shorter} overall ($0.88{\times}$).  Only Claude preserves the
Mintaka direction ($1.14{\times}$), while GPT-5.1 ($0.78{\times}$) and
Gemini ($0.85{\times}$) show shorter incorrect outputs.  The reversal is
question-type-dependent: domain-relevant questions still show longer
incorrect answers ($1.20{\times}$), but metrics-generated questions
strongly reverse ($0.62{\times}$), indicating that verbosity as a quality
signal does not generalise across domains.

\label{sec:results:confidence}
\textbf{Confidence scores are poorly calibrated across all families on both tasks.}
On Mintaka, mean confidence on incorrect answers exceeds 0.91 for every
family, with family-mean correct--incorrect gaps of only 0.015--0.022
(Figure~\ref{fig:conf-correctness}).  Disaggregating by effort level
reveals wider per-configuration gaps: GPT-5.1's gap narrows from 0.028
(low effort) to 0.007 (high), and Claude Opus~4.6's from 0.017 to 0.010,
while Gemini is invariant.  Uncertain predictions---those the rule
engine could not settle---receive noticeably lower confidence from GPT-5.1
and Claude; Gemini's uncertain confidence is indistinguishable from correct.
On FinanceBench the same pattern holds: mean confidence on incorrect
answers is 0.904 overall, with a correct--incorrect gap of only 0.028.
Gemini is the most overconfident (0.958 on incorrect), while GPT-5.1 shows
the largest discriminative gap (0.048).  Unlike Mintaka, higher reasoning
effort \emph{widens} the gap for GPT-5.1 (from 0.027 at low to 0.066 at
high), suggesting that extended reasoning improves calibration on
domain-specific content.  Across both tasks, 93\% of FinanceBench
confidence scores exceed 0.8, confirming that self-reported confidence is
not a reliable filter for incorrect answers regardless of domain.

\label{sec:results:semantic-gap}
\textbf{Semantic similarity alone does not predict cross-lingual accuracy.}
On Mintaka, all eight target languages show lower answer-level cosine
similarity than English, with a clear three-tier structure
(Table~\ref{tab:semantic-language-gaps}): Hindi is a pronounced outlier,
Arabic and Japanese form a middle tier, and the five European languages
cluster near English.  This ranking is robust across both Ada and
MPNet encoders on Mintaka (Pearson $r = 0.987$), but the two encoders
diverge on FinanceBench---German's Ada gap is $-0.004$ while its MPNet gap
is $-0.045$---further undermining similarity as a reliable signal.
On FinanceBench, answer-level gaps are an order of magnitude smaller (Ada
$\Delta$ from $-0.007$ to $-0.001$), and at the language level,
similarity is \emph{anti-correlated} with accuracy---Hindi has the lowest
answer similarity yet the highest accuracy.  Neither encoder's similarity
score correlates with item-level correctness (Pearson $r < 0.01$,
$p > 0.75$), indicating that semantic similarity alone is not a usable
proxy for cross-lingual accuracy.

\begin{table*}[t]
  \centering
  \small
  \setlength{\tabcolsep}{4pt}
  \begin{tabular}{lrrrrrr}
    \toprule
    & & \multicolumn{2}{c}{\textbf{Ada}} &
    \multicolumn{2}{c}{\textbf{MPNet}} \\
    \cmidrule(lr){3-4}\cmidrule(lr){5-6}
    \textbf{Language} & \textbf{$n$} &
    \textbf{$\Delta$} & \textbf{95\% interval} &
    \textbf{$\Delta$} & \textbf{95\% interval} \\
    \midrule
    \multicolumn{6}{l}{\emph{Task~2: Mintaka (prediction vs ground truth)}} \\
    \midrule
    Hindi      & 243 & $-0.0790$ & $[-0.0880,-0.0701]$ & $-0.1167$ & $[-0.1382,-0.0964]$ \\
    Arabic     & 240 & $-0.0323$ & $[-0.0384,-0.0263]$ & $-0.0314$ & $[-0.0470,-0.0159]$ \\
    Japanese   & 238 & $-0.0215$ & $[-0.0267,-0.0162]$ & $-0.0166$ & $[-0.0321,-0.0014]$ \\
    Portuguese & 245 & $-0.0095$ & $[-0.0128,-0.0066]$ & $-0.0108$ & $[-0.0199,-0.0021]$ \\
    Spanish    & 245 & $-0.0085$ & $[-0.0117,-0.0056]$ & $-0.0117$ & $[-0.0212,-0.0029]$ \\
    French     & 244 & $-0.0083$ & $[-0.0112,-0.0056]$ & $-0.0038$ & $[-0.0108,\phantom{-}0.0026]$ \\
    Italian    & 239 & $-0.0071$ & $[-0.0108,-0.0040]$ & $-0.0059$ & $[-0.0144,\phantom{-}0.0019]$ \\
    German     & 244 & $-0.0047$ & $[-0.0063,-0.0031]$ & $-0.0029$ & $[-0.0083,\phantom{-}0.0021]$ \\
    \midrule
    \multicolumn{6}{l}{\emph{Task~3: FinanceBench (prediction vs ground truth)}} \\
    \midrule
    Hindi      & 450 & $-0.0072$ & $[-0.0104,-0.0040]$ & $\phantom{-}0.0039$ & $[-0.0106,\phantom{-}0.0184]$ \\
    Spanish    & 450 & $-0.0013$ & $[-0.0046,\phantom{-}0.0019]$ & $-0.0005$ & $[-0.0147,\phantom{-}0.0135]$ \\
    German     & 450 & $-0.0036$ & $[-0.0073,-0.0001]$ & $-0.0445$ & $[-0.0594,-0.0305]$ \\
    \bottomrule
  \end{tabular}
  \caption{Answer semantic-similarity gaps by target language under Ada
    and MPNet.  Each $\Delta$ is target minus English cosine similarity,
    paired by item and model condition.  On Mintaka, Hindi shows the largest gap under
    both encoders; on FinanceBench, gaps are an order of magnitude smaller
    and German is the only language with a significant MPNet deficit.
    $n$ is the number of paired comparisons.}
  \label{tab:semantic-language-gaps}
\end{table*}

\label{sec:results:rq1-interaction}%
\enlargethispage{\baselineskip}
The language~$\times$~configuration interaction is negligible for GPT-5.1
and Claude Opus~4.6 (both intervals include zero) and only marginally positive
for Gemini~3.5 Flash, indicating
that provider-exposed reasoning-depth changes do not differentially
benefit non-English languages over English in entity-answer semantic
\mbox{similarity}.

\label{sec:results:citation-bias}
\textbf{We also observe that citation behaviour is surprisingly quite stable across languages, with no consistent hierarchy.}
Citation presence is near-universal (97.6--99.0\%), and citation-set
Jaccard similarity with English averages 77--79\% across all target
languages (Table~\ref{tab:citation-bias} in
Appendix~\ref{app:tables}), indicating that approximately one-fifth of
cited references differ.  The few statistically significant comparisons (after
Benjamini--Hochberg correction) show mixed directions across language,
model family, and reasoning level---no consistent language hierarchy
emerges.  Proxy precision is stable (78--79\%), while recall tends slightly lower
for several languages---most notably Italian ($-3.6$pp), Japanese
($-3.1$pp), and Hindi ($-2.9$pp), with German close behind
($-2.3$pp)---weakly suggesting that some localised conditions cite a
narrower set, though no individual comparison survives correction.  The
amount of source-selection change depends more on the model--effort
configuration than on language.

\label{sec:results:financebench}
\textbf{Cross-task reusability: which Mintaka patterns transfer to
financial QA?}
Table~\ref{tab:reusability} tests whether seven quantitative patterns
observed on Mintaka transfer to FinanceBench.  Three transfer
(model-family ranking, numerical-answer robustness, and poor confidence
calibration), two transfer partially: reasoning-depth lift is
directionally consistent but $4.0{\times}$ larger on financial content,
while the cross-lingual gap shrinks and becomes inconsistent across
families (Gemini reverses sign).  Two fail: the language-difficulty
ranking reverses, and the verbosity-as-incorrectness signal reverses on
financial content.  In both domains, self-reported confidence exceeds
0.90 on incorrect answers, making it an unreliable quality filter.

\begin{table}[t]
  \centering
  \footnotesize
  \setlength{\tabcolsep}{2pt}
  \resizebox{\columnwidth}{!}{%
  \begin{tabular}{@{}lllc@{}}
    \toprule
    \textbf{Dimension} & \textbf{Mintaka} & \textbf{FinBench} & \textbf{Xfer?} \\
    \midrule
    Lang.\ diff.\ rank     & en$>$es$>$de$>$hi & hi$>$es$>$en$>$de & No \\
    Cross-ling.\ gap       & {$+$0.9\,pp}      & {$+$0.2\,pp (mix)} & Partial \\
    Depth lift (lo$\to$hi) & {$+$0.5\,pp}      & {$+$2.0\,pp}       & Partial \\
    Model fam.\ rank       & Cl$>$GPT$>$Gem    & Cl$>$GPT$>$Gem     & Yes$^{*}$ \\
    Numer.\ robust.        & {$+$1.0\,pp}      & {$+$2.0\,pp}       & Yes \\
    Verbose$=$incorrect    & All longer ($1.95{\times}$) & Mixed ($0.88{\times}$) & No \\
    Poor calibration       & gap 0.015--0.022  & gap 0.028           & Yes \\
    \bottomrule
  \end{tabular}}
  \caption{Cross-task reusability: do Mintaka patterns predict FinanceBench?
    Language ranking and the verbosity signal do not transfer;
    cross-lingual gap magnitude and reasoning-depth lift transfer only
    partially; model ranking, numerical robustness, and poor confidence
    calibration transfer.  $^{\ast}$Overall ranking is preserved, but at high
    effort GPT-5.1 leads on FinanceBench while Claude leads on Mintaka.
    Gaps are mean en$-$target accuracy differences across the three
    model families at high effort, computed over the three non-English
    languages shared between benchmarks (Hindi, Spanish, German).}
  \label{tab:reusability}
\end{table}


\section{Discussion}
\label{sec:discussion}

Our results converge on several cross-cutting findings that inform multilingual
LLM deployment, particularly for financial institutions considering
cross-lingual pipeline reuse.

\paragraph{Cross-lingual reusability.}
Across all three tasks, the evidence supports qualified optimism.
Accuracy drops at most 3.3pp on Task~2, Gemini is near-ceiling across
prompt languages on Task~1 ($\geq$95.2\%), and FinanceBench
per-language means span only 5.8pp.  However, reusability is
model-dependent---Claude degrades with non-English prompts on
Task~1---and the patterns that fail or transfer only partially
(Table~\ref{tab:reusability}, rows~1--3) mean that per-language
validation on the target domain is essential before deployment.

\paragraph{Limitations.}
Several limitations temper these conclusions.
First, the pinned evidence corpus was captured in English and translated
using GPT-4.1; observed cross-lingual gaps therefore conflate reasoning
degradation with translation-quality variation, and native-language
retrieval remains untested.
Second, proxy precision and recall rely on alias-coattributed references, not
human relevance judgments.
Third, the semantic-similarity diagnostic uses a single embedding model and
no calibrated threshold; it is not a substitute for factual accuracy.
Fourth, the reasoning-depth axis is not commensurable across families
(GPT uses a graded ladder, Claude an on/off switch, Gemini a narrow range),
so cross-family depth comparisons should be read with caution.
Finally, the 400-item Mintaka sample---while stratified---may not capture
all difficulty profiles present in full-scale deployments.  The 50-item
FinanceBench pilot addresses the domain gap and yields a substantive
finding (Hindi's ranking reverses between general-knowledge and financial
QA), but the pilot remains small and covers only four of the nine
languages; larger-scale replication is needed to confirm the pattern.


\section{Conclusion}
\label{sec:conclusion}

We presented a systematic multilingual evaluation of three major LLM families
across three tasks---financial cheque extraction,
cross-lingual question answering on a general-knowledge benchmark, and a
domain-validation pilot on financial QA over SEC~10-K filings---spanning up
to nine languages and nine model--effort configurations.

Addressing our three reusability dimensions: \emph{accuracy reusability}
is broadly confirmed---non-English prompts can match or exceed English
on extraction and QA, though the language-difficulty ranking is
domain-dependent (\textbf{RQ1}).  \emph{Effort reusability} holds within
domains: the direction of reasoning-effort effects transfers across
languages, but their magnitude---and, for some families, their
direction---does not transfer reliably across domains (\textbf{RQ2}).
\emph{Behavioural reusability} is stable: citation-set overlap with
English averages ${\sim}78\%$ Jaccard, with no single language
consistently worse (\textbf{RQ3}).
The FinanceBench domain-validation pilot confirms that model-family
dominance and small cross-lingual gaps extend to financial content
(79.9\% overall, 5.8pp best-to-worst spread), while revealing that the
language-difficulty ranking is domain-dependent: Hindi, the weakest
language on general-knowledge QA, is the strongest on financial QA.
Additionally, output token cost is family-dependent: Claude's
cross-language range exceeds $20\%$ (Hindi is the most expensive
language), while GPT is nearly language-neutral, making model selection
the dominant cost driver for multilingual deployments.
More broadly, the modest cross-lingual accuracy and citation-behaviour
gaps suggest that English-developed pipelines can often be reused across
languages through translation, but per-language validation on the target
domain remains necessary---semantic-similarity screening does not
reliably predict accuracy across domains.

Future work should extend this analysis in three directions:
(i)~native-language retrieval to disentangle search quality from
reasoning quality, (ii)~LLM-judged citation support precision and
completeness to move beyond proxy reference labels, and
(iii)~larger-scale financial evaluations across more languages to
test whether the Hindi reversal and small cross-lingual gaps observed
in the 50-item pilot hold at scale.


\bibliographystyle{plainnat}
\bibliography{references}

\begin{thebibliography}{35}
\providecommand{\natexlab}[1]{#1}
\providecommand{\url}[1]{\texttt{#1}}
\expandafter\ifx\csname urlstyle\endcsname\relax
  \providecommand{\doi}[1]{doi: #1}\else
  \providecommand{\doi}{doi: \begingroup \urlstyle{rm}\Url}\fi

\bibitem[{Anthropic}(2025)]{anthropic2025sonnet45}
{Anthropic}.
\newblock {Claude Sonnet 4.5 System Card}.
\newblock System Card, September 2025.
\newblock
  \href{https://www-cdn.anthropic.com/963373e433e489a87a10c823c52a0a013e9172dd.pdf}{Anthropic
  system card}.

\bibitem[{Anthropic}(2026)]{anthropic2026opus46}
{Anthropic}.
\newblock {Claude Opus 4.6 System Card}.
\newblock System Card, February 2026.
\newblock \url{https://www.anthropic.com/claude-opus-4-6-system-card}.

\bibitem[Chen et~al.(2025{\natexlab{a}})]{chen2025pairwisebias}
Guiming Chen et~al.
\newblock Bias in pairwise {LLM} evaluation.
\newblock In \emph{Proceedings of BlackboxNLP 2025}, 2025{\natexlab{a}}.
\newblock \url{https://aclanthology.org/2025.blackboxnlp-1.5/}.

\bibitem[Chen et~al.(2025{\natexlab{b}})]{chen2025languageofthought}
Yifan Chen et~al.
\newblock Language of thought shapes output diversity in large language models.
\newblock In \emph{arXiv preprint arXiv:2601.11227}, 2025{\natexlab{b}}.
\newblock \url{https://arxiv.org/abs/2601.11227}.

\bibitem[Deng et~al.(2026)]{deng2026thoughtsdepth}
Jiayu Deng et~al.
\newblock Do thoughts depth affect multilingual reasoning?
\newblock In \emph{Proceedings of the Workshop on Multilingual Large Language
  Models (MeLLM)}, 2026.
\newblock \url{https://aclanthology.org/2026.mellm-1.14.pdf}.

\bibitem[Gao et~al.(2026)]{gao2026languagelatent}
Yue Gao et~al.
\newblock Language as a latent variable for reasoning optimization.
\newblock In \emph{arXiv preprint arXiv:2604.21593}, 2026.
\newblock \url{https://arxiv.org/abs/2604.21593}.

\bibitem[{Google DeepMind}(2026)]{google2026gemini35}
{Google DeepMind}.
\newblock {Gemini 3.5 Flash Model Card}.
\newblock Model Card, May 2026.
\newblock \url{https://deepmind.google/models/model-cards/gemini-3-5-flash/}.

\bibitem[Hada et~al.(2025)]{hada2025multilingualjudge}
Rishav Hada et~al.
\newblock Multilingual {LLM}-as-a-judge: Evaluating across languages.
\newblock In \emph{Proceedings of EMNLP 2025}, 2025.
\newblock \url{https://arxiv.org/abs/2505.12201}.

\bibitem[Haldar and Hockenmaier(2025)]{haldar2025ratingroulette}
Aparajita Haldar and Julia Hockenmaier.
\newblock Rating roulette: Self-inconsistency in {LLM}-as-a-judge frameworks.
\newblock In \emph{Findings of the Association for Computational Linguistics:
  EMNLP 2025}, 2025.
\newblock \url{https://arxiv.org/abs/2510.27106}.

\bibitem[Islam et~al.(2023)Islam, Nair, Maguire, Finn, Giorgi, and
  Keenan]{islam2023financebench}
Pranab Islam, Anand Nair, Eoin Maguire, Aidan Finn, John Giorgi, and Olga
  Keenan.
\newblock {FinanceBench}: A new benchmark for financial question answering.
\newblock \emph{arXiv preprint arXiv:2311.11944}, 2023.
\newblock \url{https://arxiv.org/abs/2311.11944}.

\bibitem[Kim et~al.(2022)Kim, Hong, Yim, Nam, Park, Yim, Hwang, Yun, Han, and
  Park]{kim2022donut}
Geewook Kim, Teakgyu Hong, Moonbin Yim, JeongYeon Nam, Jinyoung Park, Jinyeong
  Yim, Wonseok Hwang, Sangdoo Yun, Dongyoon Han, and Seunghyun Park.
\newblock {OCR}-free document understanding transformer.
\newblock \emph{arXiv preprint arXiv:2111.15664}, 2022.
\newblock \url{https://arxiv.org/abs/2111.15664}.

\bibitem[Krippendorff(2011)]{krippendorff2011alpha}
Klaus Krippendorff.
\newblock Computing {Krippendorff's} alpha-reliability.
\newblock \emph{Communication Methods and Measures}, 2011.

\bibitem[Lee et~al.(2026)]{lee2026fairnessfluency}
Dongryeol Lee et~al.
\newblock Fairness or fluency? an investigation into language bias of pairwise
  {LLM}-as-a-judge.
\newblock In \emph{arXiv preprint arXiv:2601.13649}, 2026.
\newblock \url{https://arxiv.org/abs/2601.13649}.

\bibitem[Lu et~al.(2026)Lu, Zeng, Qi, Yu, Zhao, Yang, Yanaka, Yokoya, and
  Xuan]{lu2026beyondmonolingual}
Yuheng Lu, Qingcheng Zeng, Heli Qi, Puxuan Yu, Fuheng Zhao, Rui Yang, Hitomi
  Yanaka, Naoto Yokoya, and Weihao Xuan.
\newblock Beyond monolingual deep research: Evaluating agents and retrievers
  with cross-lingual {BrowseComp-Plus}.
\newblock In \emph{arXiv preprint arXiv:2606.15345}, 2026.
\newblock \url{https://arxiv.org/abs/2606.15345}.

\bibitem[Ma et~al.(2024)]{ma2024mmlongbench}
Yubo Ma et~al.
\newblock {MMLONGBENCH-DOC}: Benchmarking long-context document understanding
  with visualizations.
\newblock In \emph{Advances in Neural Information Processing Systems (NeurIPS),
  Datasets and Benchmarks Track}, 2024.
\newblock \url{https://arxiv.org/abs/2407.01523}.

\bibitem[Mondshine et~al.(2025)Mondshine, Paz-Argaman, and
  Tsarfaty]{mondshine2025beyondenglish}
Itai Mondshine, Tzuf Paz-Argaman, and Reut Tsarfaty.
\newblock Beyond {English}: The impact of prompt translation strategies across
  languages and tasks in multilingual {LLMs}.
\newblock \emph{arXiv preprint arXiv:2502.09331}, 2025.
\newblock \url{https://arxiv.org/abs/2502.09331}.

\bibitem[{OpenAI}(2022)]{openai2022ada002}
{OpenAI}.
\newblock {New and Improved Embedding Model}, December 2022.
\newblock \url{https://openai.com/index/new-and-improved-embedding-model/}.

\bibitem[{OpenAI}(2025{\natexlab{a}})]{openai2025gpt41}
{OpenAI}.
\newblock {Introducing GPT-4.1 in the API}, April 2025{\natexlab{a}}.
\newblock \url{https://openai.com/index/gpt-4-1/}.

\bibitem[{OpenAI}(2025{\natexlab{b}})]{openai2025gpt51}
{OpenAI}.
\newblock {GPT-5.1 Instant and GPT-5.1 Thinking System Card Addendum}.
\newblock System Card, November 2025{\natexlab{b}}.
\newblock \url{https://openai.com/index/gpt-5-system-card-addendum-gpt-5-1/}.

\bibitem[{OpenAI}(2025{\natexlab{c}})]{openai2025gpt52}
{OpenAI}.
\newblock {Update to GPT-5 System Card: GPT-5.2}.
\newblock System Card, December 2025{\natexlab{c}}.
\newblock \url{https://openai.com/index/gpt-5-system-card-update-gpt-5-2/}.

\bibitem[Peng et~al.(2026)Peng, Han, Xie, et~al.]{peng2026multifinben}
Xueqing Peng, Weiguang Han, Qianqian Xie, et~al.
\newblock {MultiFinBen}: Benchmarking large language models for multilingual
  and multimodal financial application.
\newblock In \emph{Proceedings of the 64th Annual Meeting of the Association
  for Computational Linguistics (ACL)}, 2026.
\newblock \url{https://arxiv.org/abs/2506.14028}.

\bibitem[Pfitzmann et~al.(2022)Pfitzmann, Auer, Dolfi, Nassar, and
  Staar]{pfitzmann2022doclaynet}
Birgit Pfitzmann, Christoph Auer, Michele Dolfi, Ahmed~S. Nassar, and Peter
  W.~J. Staar.
\newblock {DocLayNet}: A large human-annotated dataset for document-layout
  segmentation.
\newblock In \emph{Proceedings of the 28th ACM SIGKDD Conference on Knowledge
  Discovery and Data Mining (KDD)}, pages 3743--3751, 2022.
\newblock \url{https://arxiv.org/abs/2206.01062}.

\bibitem[Qin et~al.(2025)]{qin2025multilingualcot}
Libo Qin et~al.
\newblock A comprehensive evaluation of multilingual chain-of-thought
  reasoning: Performance, consistency, and faithfulness across languages.
\newblock In \emph{arXiv preprint arXiv:2510.09555}, 2025.
\newblock \url{https://arxiv.org/abs/2510.09555}.

\bibitem[Sapenov and Savelka(2026)]{sapenov2026mmpisabench}
Yerzhan Sapenov and Jaromir Savelka.
\newblock {mmPISA-bench}: Do {LLMs} reason equally well across 43 languages?
\newblock In \emph{Proceedings of the 1st Workshop on Multilinguality in the
  Era of Large Language Models (MeLLM)}, pages 170--180, 2026.
\newblock \url{https://arxiv.org/abs/2606.07069}.

\bibitem[Sen et~al.(2022)Sen, Aji, and Saffari]{sen2022mintaka}
Priyanka Sen, Alham~Fikri Aji, and Amir Saffari.
\newblock Mintaka: A complex, natural, and multilingual dataset for end-to-end
  question answering.
\newblock In \emph{Proceedings of the 29th International Conference on
  Computational Linguistics (COLING)}, 2022.

\bibitem[Shah et~al.(2022)Shah, Kuhn, Vyas, and Bansal]{shah2022flue}
Raj~Sanjay Shah, Kunal Kuhn, Ravi Vyas, and Mohit Bansal.
\newblock When {FLUE} meets {FLANG}: Benchmarks and large pre-trained language
  model for financial domain.
\newblock In \emph{Proceedings of EMNLP 2022}, 2022.
\newblock \url{https://arxiv.org/abs/2211.00083}.

\bibitem[Song et~al.(2020)Song, Tan, Qin, Lu, and Liu]{song2020mpnet}
Kaitao Song, Xu~Tan, Tao Qin, Jianfeng Lu, and Tie-Yan Liu.
\newblock {MPNet: Masked and Permuted Pre-training for Language Understanding}.
\newblock In \emph{Advances in Neural Information Processing Systems},
  volume~33, pages 16857--16867, 2020.
\newblock
  \href{https://proceedings.neurips.cc/paper/2020/hash/c3a690be93aa602ee2dc0ccab5b7b67e-Abstract.html}{NeurIPS
  proceedings}.

\bibitem[Thellmann et~al.(2026)Thellmann, Stadler, F{\"a}rber, and
  Lehmann]{thellmann2026translationerrors}
Klaudia Thellmann, Bernhard Stadler, Michael F{\"a}rber, and Jens Lehmann.
\newblock Quantifying the impact of translation errors on multilingual {LLM}
  evaluation.
\newblock In \emph{Proceedings of the 64th Annual Meeting of the Association
  for Computational Linguistics (Volume 1: Long Papers)}, pages 41311--41334,
  2026.
\newblock \doi{10.18653/v1/2026.acl-long.1916}.
\newblock URL \url{https://aclanthology.org/2026.acl-long.1916/}.

\bibitem[Verma(2024)]{verma2024chequeVQA}
Aniket Verma.
\newblock Handwritten cheque {VQA} dataset.
\newblock Hugging Face Datasets, 2024.
\newblock
  \url{https://huggingface.co/datasets/aniketVerma07/handwritten_cheque_vqa}.

\bibitem[Wang et~al.(2026)Wang, He, Chen, Yehudai, Liu, Ying, Shmueli-Scheuer,
  and Cohan]{wang2026reflect}
Leyao Wang, Yanan He, Peng Chen, Asaf Yehudai, Yixin Liu, Rex Ying, Michal
  Shmueli-Scheuer, and Arman Cohan.
\newblock Time to {REFLECT}: Can we trust {LLM} judges for evidence-based
  research agents?
\newblock In \emph{arXiv preprint arXiv:2605.19196}, 2026.
\newblock \url{https://arxiv.org/abs/2605.19196}.

\bibitem[Wang et~al.(2025)]{wang2025explang}
Yifei Wang et~al.
\newblock {ExpLang}: Improved exploration and exploitation in {LLM} reasoning
  with on-policy thinking language selection.
\newblock In \emph{arXiv preprint arXiv:2602.21887}, 2025.
\newblock \url{https://arxiv.org/abs/2602.21887}.

\bibitem[Wu et~al.(2023)Wu, Irsoy, Lu, Dabravolski, Dredze, Gehrmann, Kambadur,
  Rosenberg, and Mann]{wu2023bloomberggpt}
Shijie Wu, Ozan Irsoy, Steven Lu, Vadim Dabravolski, Mark Dredze, Sebastian
  Gehrmann, Prabhanjan Kambadur, David Rosenberg, and Gideon Mann.
\newblock {BloombergGPT}: A large language model for finance.
\newblock \emph{arXiv preprint arXiv:2303.17564}, 2023.
\newblock \url{https://arxiv.org/abs/2303.17564}.

\bibitem[Xie et~al.(2024)Xie, Han, Chen, Xiang, Zhang, He, Xiao, Li, Dai, Feng,
  et~al.]{xie2024finben}
Qianqian Xie, Weiguang Han, Zhengyu Chen, Ruoyu Xiang, Xiao Zhang, Yueru He,
  Mengxi Xiao, Dong Li, Yongfu Dai, Duanyu Feng, et~al.
\newblock {FinBen}: A holistic financial benchmark for large language models.
\newblock In \emph{Proceedings of NeurIPS Datasets and Benchmarks Track}, 2024.
\newblock \url{https://arxiv.org/abs/2402.12659}.

\bibitem[Ye et~al.(2024)]{ye2024prepair}
Xi~Ye et~al.
\newblock {PRePair}: Pointwise audit before pairwise comparison for {LLM}
  evaluation.
\newblock In \emph{arXiv preprint arXiv:2406.07791}, 2024.
\newblock \url{https://arxiv.org/abs/2406.07791}.

\bibitem[Zhao et~al.(2024)Zhao, Chen, Huang, Zhang, Warner, Chen, Tang, Ma, and
  Nan]{zhao2024gencheck}
Fei Zhao, Jiawen Chen, Bin Huang, Chengcui Zhang, Gary Warner, Rushi Chen,
  Shaorou Tang, Yuanfei Ma, and Zixi Nan.
\newblock {GenCheck}: A {LoRA}-adapted multimodal large language model for
  check analysis.
\newblock In \emph{7th IEEE International Conference on Multimedia Information
  Processing and Retrieval (MIPR)}, pages 88--94, 2024.
\newblock \url{https://doi.org/10.1109/mipr62202.2024.00021}.

\end{thebibliography}


\appendix

\section{Supplementary Tables}
\label{app:tables}

\begin{table}[h]
  \centering
  \scriptsize
  \setlength{\tabcolsep}{2.5pt}
  \begin{tabular}{lrrrrrr}
    \toprule
    \textbf{Lang.} & \textbf{Cited} & \textbf{Cov.} &
    \textbf{Blanket} & \textbf{Prec.} & \textbf{Recall} &
    \textbf{Jaccard} \\
    \midrule
    en & 97.6\% & 64.0\% & 44.7\% & 78.2\% & 69.5\% & --- \\
    ar & 98.7  & 63.1  & 43.9  & 79.0  & 68.0  & 78.2\% \\
    de & 98.3  & 62.2  & 43.7  & 78.6  & 67.2  & 78.8  \\
    es & 98.6  & 64.0  & 46.1  & 78.2  & 68.8  & 79.1  \\
    fr & 98.8  & 63.7  & 45.8  & 78.1  & 68.5  & 79.0  \\
    hi & 99.0  & 61.3  & 41.5  & 79.5  & 66.6  & 78.2  \\
    it & 98.0  & 61.2  & 43.2  & 79.3  & 65.9  & 77.5  \\
    ja & 98.1  & 62.3  & 45.2  & 79.0  & 66.4  & 77.2  \\
    pt & 98.9  & 63.8  & 46.1  & 78.8  & 68.6  & 79.3  \\
    \bottomrule
  \end{tabular}
  \caption{Citation and source-selection metrics averaged over the nine
    model--effort configurations.  \emph{Cited}: answers with $\geq$1
    citation; \emph{Cov.}: share of available references cited;
    \emph{Blanket}: citing every reference; \emph{Prec./Recall}: proxy
    alias-coattributed precision and recall (boolean items excluded);
    \emph{Jaccard}: citation-set overlap with English.  All target-language
    comparisons are localised-condition effects, not pure language effects.}
  \label{tab:citation-bias}
\end{table}

\section{Supplementary Figures}
\label{app:figures}

\begin{figure*}[ht]
  \centering
  \begin{subfigure}[t]{\textwidth}
    \centering
    \begingroup
    \setlength{\unitlength}{1pt}
    \setlength{\fboxsep}{1pt}
    \begin{picture}(469.755,173.864)
      \put(0,0){\includegraphics[width=\textwidth]
        {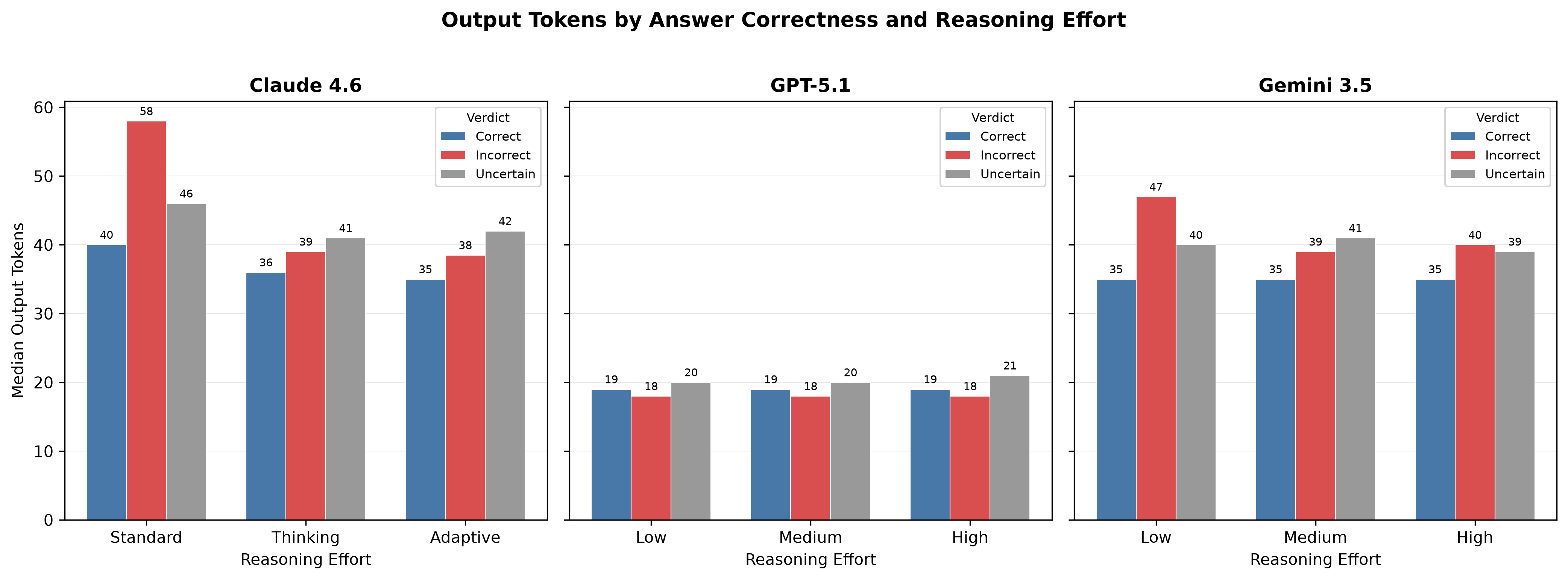}}
      \put(40,151){\colorbox{white}{\rule[-6pt]{0pt}{13pt}\makebox[100pt][c]{\fontsize{6}{7}\selectfont\bfseries Claude Opus 4.6}}}
      \put(340,151){\colorbox{white}{\rule[-6pt]{0pt}{13pt}\makebox[100pt][c]{\fontsize{6}{7}\selectfont\bfseries Gemini 3.5 Flash}}}
    \end{picture}
    \endgroup
    \caption{Median output tokens split by answer correctness and reasoning
      effort, counted with each family's native tokenizer.  Claude Opus~4.6 in
      standard mode shows the largest median gap (58 vs.\ 40 for incorrect
      vs.\ correct); GPT-5.1 incorrect answers are shorter at the median
      despite a higher mean, reflecting a heavy right tail.  Gemini~3.5 Flash
      shows a consistent but modest gap across all effort levels.}
    \label{fig:outlen-correctness}
  \end{subfigure}

  \vspace{0.8em}

  \begin{subfigure}[t]{\textwidth}
    \centering
    \includegraphics[width=\textwidth]
      {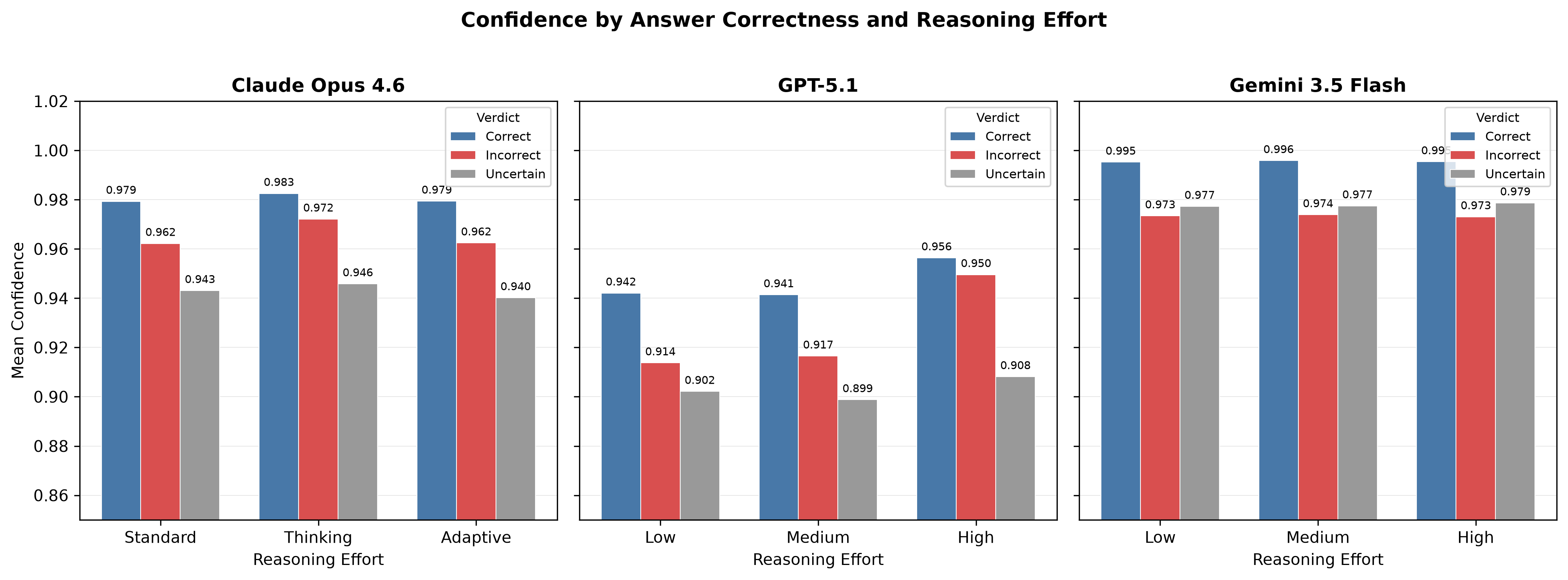}
    \caption{Mean self-reported confidence by answer correctness and
      reasoning effort.  All three families are poorly calibrated:
      incorrect answers receive mean confidence above~0.91.  GPT-5.1 and
      Claude Opus~4.6 show noticeably lower confidence on uncertain items,
      while Gemini~3.5 Flash's uncertain-answer confidence is nearly
      indistinguishable from correct.}
    \label{fig:conf-correctness}
  \end{subfigure}

  \caption{Output characteristics split by answer correctness and reasoning
    effort across three model families.}
  \label{fig:correctness-combined}
\end{figure*}

\end{document}